%% file: paper.tex
\documentclass[runningheads]{llncs}
\usepackage{graphicx}
\usepackage{tikz}
\usepackage{comment} 
\usepackage{amsmath,amssymb}
\usepackage{color}
\usepackage{epsfig}
\usepackage{graphicx}
\usepackage{cases}
\usepackage{gensymb}
\usepackage{subfigure}
\usepackage{booktabs}
\usepackage{bm}
\usepackage{multirow}
\usepackage{url}

\begin{document}
\pagestyle{headings}
\mainmatter

\title{PIoU Loss: Towards Accurate Oriented \\ Object Detection in Complex Environments}


\titlerunning{PIoU Loss}
%
\author{Zhiming Chen\inst{1,2} \and Kean Chen\inst{2} \and Weiyao Lin\inst{2\star} \and John See\inst{3} \and \\ Hui Yu\inst{1} \and Yan Ke\inst{1} \and Cong Yang\inst{1}\thanks{Corresponding author: \email{cong.yang@clobotics.com}, \email{wylin@sjtu.edu.cn}}\\}
\authorrunning{Z. Chen et al.}
%
\institute{Clobotics, China \and Department of Electronic Engineering, Shanghai Jiao Tong University, China \and Faculty of Computing and Informatics, Multimedia University, Malaysia}

\maketitle

\begin{abstract}
Object detection using an oriented bounding box (OBB) can better target rotated objects by reducing the overlap with background areas. Existing OBB approaches are mostly built on horizontal bounding box detectors by introducing an additional angle dimension optimized by a distance loss. However, as the distance loss only minimizes the angle error of the OBB and that it loosely correlates to the IoU, it is insensitive to objects with high aspect ratios. Therefore, a novel loss, Pixels-IoU (PIoU) Loss, is formulated to exploit both the angle and IoU for accurate OBB regression. The PIoU loss is derived from IoU metric with a pixel-wise form, which is simple and suitable for both horizontal and oriented bounding box. To demonstrate its effectiveness, we evaluate the PIoU loss on both anchor-based and anchor-free frameworks. The experimental results show that PIoU loss can dramatically improve the performance of OBB detectors, particularly on objects with high aspect ratios and complex backgrounds. Besides, previous evaluation datasets did not include scenarios where the objects have high aspect ratios, hence a new dataset, Retail50K, is introduced to encourage the community to adapt OBB detectors for more complex environments.
\end{abstract}

\input{introduction}
\input{related}
\input{piou}
\input{retail50k}
\input{experiment}
\input{conclusion}

\section*{Acknowledgements}
The paper is supported in part by the following grants:  China Major Project for New Generation of AI Grant (No.2018AAA0100400), National Natural Science Foundation of China (No. 61971277). The work is also supported by funding from Clobotics under the Joint Research Program of Smart Retail. 

%
%
\bibliographystyle{splncs04}
\bibliography{cong}
\end{document}

%% file: introduction.tex
\section{Introduction}
\label{intro}
Object detection is a fundamental task in computer vision and many detectors~\cite{Ren2015FRT,Liu2016SSS,Lin2017FLF,Law2018CDO} using convolutional neural networks have been proposed in recent years. In spite of their state-of-the-art performance, those detectors have inherent limitations on rotated and densely crowded objects. For example, bounding boxes (BB) of a rotated or perspective-transformed objects usually contain a significant amount of background that could mislead the classifiers. When bounding boxes have high overlapping areas, it is difficult to separate the densely crowded objects. Because of these limitations, researchers have extended existing detectors with oriented bounding boxes (OBB). In particular, as opposed to the BB which is denoted by $(c_{x}, c_{y}, w, h)$, an OBB is composed by $(c_{x}, c_{y}, w, h, \theta)$ where $(c_{x}, c_{y})$, $(w, h)$ and $\theta$ are the center point, size and rotation of an OBB, respectively. As a result, OBBs can compactly enclose the target object so that rotated and densely crowded objects can be better detected and classified.
\begin{figure}[t!]
  \centering
    \includegraphics[width=1\linewidth]{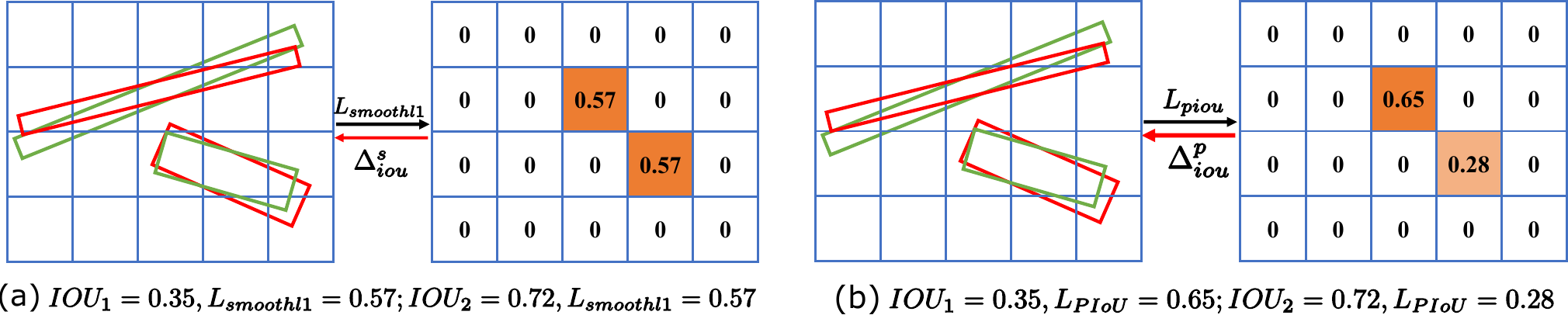}
  \caption{Comparison between PIoU and SmoothL1~\cite{Ren2015FRT} losses. (a) Loss values between IoU and SmoothL1 are totally different while their SmoothL1 loss values are the same. (b) The proposed PIoU loss is consistent and correlated with IoU.}
\label{fig:intro:ps}
\end{figure}

Existing OBB-based approaches are mostly built on anchor-based frameworks by introducing an additional angle dimension optimized by a distance loss~\cite{Liu2017LAR,Li2018MRB,Liao2018RSR,Jian2019LRT,Yang2019RRS,Yang2019STM} on the parameter tuple $(c_{x}, c_{y}, w, h, \theta)$. While OBB has been primarily used for simple rotated target detection in aerial images~\cite{Li2018MRB,Zhu2015ORO,Razakarivony2016VDI,Liu2016SRB,Liu2015FMV,Benedek2012BDM,Xia2018DAS}, the detection performance in more complex and close-up environments is limited. One of the reasons is that the distance loss in those approaches, \emph{e.g.} SmoothL1 Loss~\cite{Ren2015FRT}, mainly focus on minimizing the angle error rather than global IoU. As a result, it is insensitive to targets with high aspect ratios. An intuitive explanation is that object parts far from the center $(c_{x}, c_{y})$ are not properly enclosed even though the angle distance may be small. For example, \cite{Liao2018RSR,Jian2019LRT} employ a regression branch to extract rotation-sensitive features and thereby the angle error of the OBB can be modelled in using a transformer. However, as shown in Figure~\ref{fig:intro:ps}(a), the IoU of predicted boxes (green) and that of the ground truth (red) are very different while their losses are the same.

To solve the problem above, we introduce a novel loss function, named \textit{Pixels-IoU (PIoU) Loss}, to increase both the angle and IoU accuracy for OBB regression. In particular, as shown in Figure~\ref{fig:intro:ps}(b), the PIoU loss directly reflects the IoU and its local optimum compared to standard distance loss. The rationale behind this is that the IoU loss normally achieves better performance than the distance loss~\cite{Yu2016UAA,Rezatofighi2019GIO}. However, the IoU calculation between OBBs is more complex than BBs since the shape of intersecting OBBs could be any polygon of less than eight sides. For this reason, the PIoU, a continuous and derivable function, is proposed to jointly correlate the five parameters of OBB for checking the position (inside or outside IoU) and the contribution of each pixel. The PIoU loss can be easily calculated by accumulating the contribution of interior overlapping pixels. To demonstrate its effectiveness, the PIoU loss is evaluated on both anchor-based and anchor-free frameworks in the experiments.

To overcome the limitations of existing OBB-based approaches, we encourage the community to adopt more robust OBB detectors in a shift from conventional aerial imagery to more complex domains. We collected a new benchmark dataset, \emph{Retail50K}, to reflect the challenges of detecting oriented targets with high aspect ratios, heavy occlusions, and complex backgrounds. Experiments show that the proposed frameworks with PIoU loss not only have promising performances on aerial images, but they can also effectively handle new challenges in Retail50K.

The contributions of this work are summarized as follows: (1) We propose a novel loss function, PIoU loss, to improve the performance of oriented object detection in highly challenging conditions such as high aspect ratios and complex backgrounds. (2) We introduce a new dataset, Retail50K, to spur the computer vision community towards innovating and adapting existing OBB detectors to cope with more complex environments. (3) Our experiments demonstrate that the proposed PIoU loss can effectively improve the performances for both anchor-based and anchor-free OBB detectors in different datasets.

%% file: related.tex
\section{Related Work}

\subsection{Oriented Object Detectors}
Existing oriented object detectors are mostly extended from generic horizontal bounding box detectors by introducing an additional angle dimension. For instance, \cite{Liu2017LAR} presented a rotation-invariant detector based on one-stage SSD~\cite{Liu2016SSS}. \cite{Li2018MRB} introduced a rotated detector based on two-stage Faster RCNN~\cite{Ren2015FRT}. \cite{Jian2019LRT} designed an RoI transformer to learn the transformation from BB to OBB and thereafter, the rotation-invariant features are extracted. \cite{He2015OOP} formulated a generative probabilistic model to extract OBB proposals. For each proposal, the location, size and orientation are determined by searching the local maximum likelihood. Other possible ways of extracting OBB include, fitting detected masks~\cite{Chen2019FVO,He2017MRC} and regressing OBB with anchor-free models~\cite{Zhou2019OAP}, two new concepts in literature. While these approaches have promising performance on aerial images, they are not well-suited for oriented objects with high aspect ratios and complex environments. For this reason, we hypothesize that a new kind of loss is necessary to obtain improvements under challenging conditions. 
For the purpose of comparative evaluation, we implement both anchor-based and anchor-free frameworks as baselines in our experiments. We later show how these models, when equipped with PIoU Loss, can yield better results in both retail and aerial data.

\subsection{Regression Losses}
For bounding box regression, actively used loss functions are Mean Square Error~\cite{Mood1974ITT} (MSE, L2 loss, the sum of squared distances between target and predicted variables), Mean Absolute Error~\cite{Willmott2005AOT} (MAE, L1 loss, the sum of absolute differences between target and predicted variables), Quantile Loss~\cite{Cannon2011QRN} (an extension of MAE, predicting an interval instead of only point predictions), Huber Loss~\cite{Huber1964REO} (basically absolute error, which becomes quadratic when error is small) and Log-Cosh Loss (the logarithm of the hyperbolic cosine of the prediction error)~\cite{Muller2004OTC}. In practise, losses in common used detectors~\cite{Redmon2016YOL,Liu2016SSS,Ren2015FRT} are extended from the base functions above. However, we can not directly use them since there is an additional angle dimension involved in the OBB descriptor. 

Besides the base functions, there have been several works that introduce IoU losses for horizontal bounding box. For instance, \cite{Yu2016UAA} propose an IoU loss which regresses the four bounds of a predicted box as a whole unit. \cite{Rezatofighi2019GIO} extends the idea of~\cite{Yu2016UAA} by introducing a Generalized Intersection over Union loss (GIoU loss) for bounding box regression. The main purpose of GIoU is to get rid of the case that two polygons do not have an intersection. \cite{Smit2018LPI} introduce a novel bounding box regression loss based on a set of IoU upper bounds. However, when using oriented bounding box, those approaches become much more complicated thus are hard to implement, while the proposed PIoU loss is much simpler and suitable for both horizontal and oriented box. It should be noted that the proposed PIoU loss is different from~\cite{Zhou2019ILF} in which the IoU is computed based on axis alignment and polygon intersection, our method is more straightforward, i.e. IoU is calculated directly by accumulating the contribution of interior overlapping pixels. Moreover, the proposed PIoU loss is also different from Mask Loss in Mask RCNN~\cite{He2017MRC}. Mask loss is calculated by the average binary cross-entropy with per-pixel sigmoid (also called Sigmoid Cross-Entropy Loss). Different from it, our proposed loss is calculated based on positive IoU to preserve intersection and union areas between two boxes. In each area, the contribution of pixels are modeled and accumulated depending on their spatial information. Thus, PIoU loss is more general and sensitive to OBB overlaps.

%% file: piou.tex
\section{Pixels-IoU (PIoU) Loss}
\label{approach}
\begin{figure*}[t]
  \centering
    \includegraphics[width=1\linewidth]{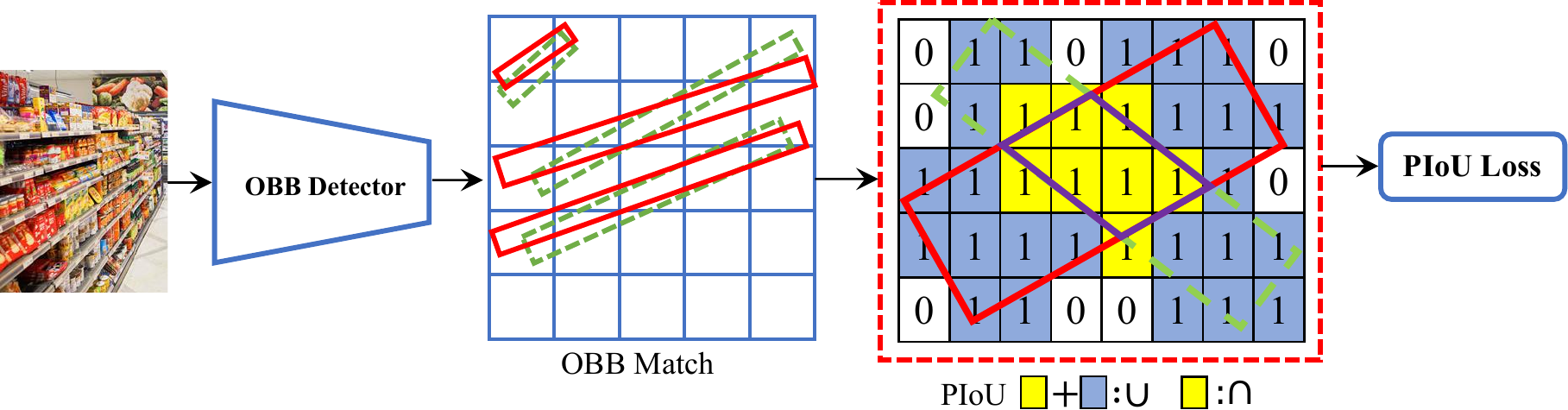}
  \vspace{-2mm}
  \caption{Our proposed PIoU is a general concept that is applicable to most OBB-based frameworks. All possible predicted (green) and g/t (red) OBB pairs are matched to compute their PIoU. Building on that, the final PIoU loss is calculated using Eq.~\ref{equ:model:finalloss}.}
  \vspace{-2mm}
\label{fig:model:general_framework}
\end{figure*}

\begin{figure}[t]
    \centering
        \subfigure[]{
        \label{fig:box_model}
        \includegraphics[width=0.35\hsize]{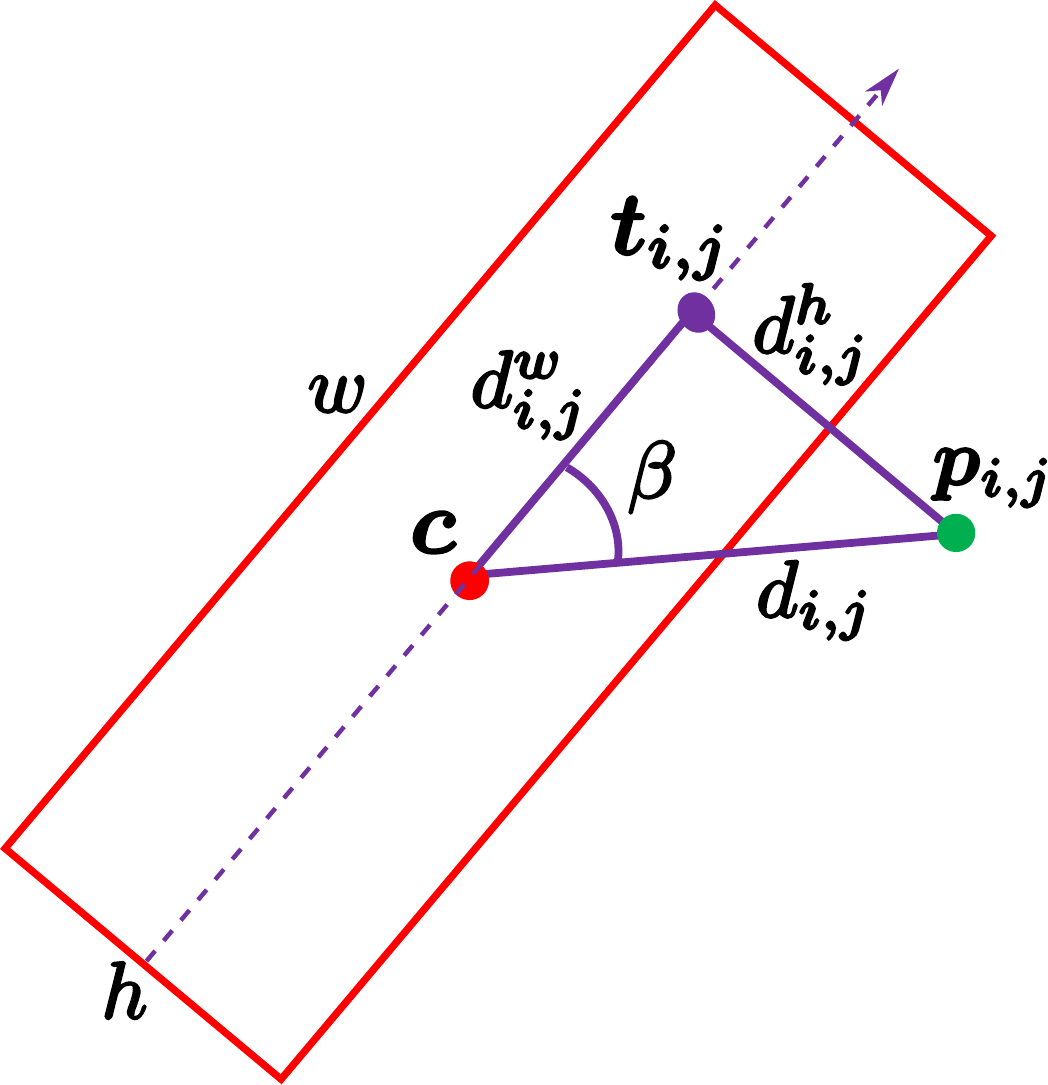}\;}
        \hspace{5mm}
        \subfigure[]{
        \label{fig:pixel_au}
        \includegraphics[width=0.45\hsize]{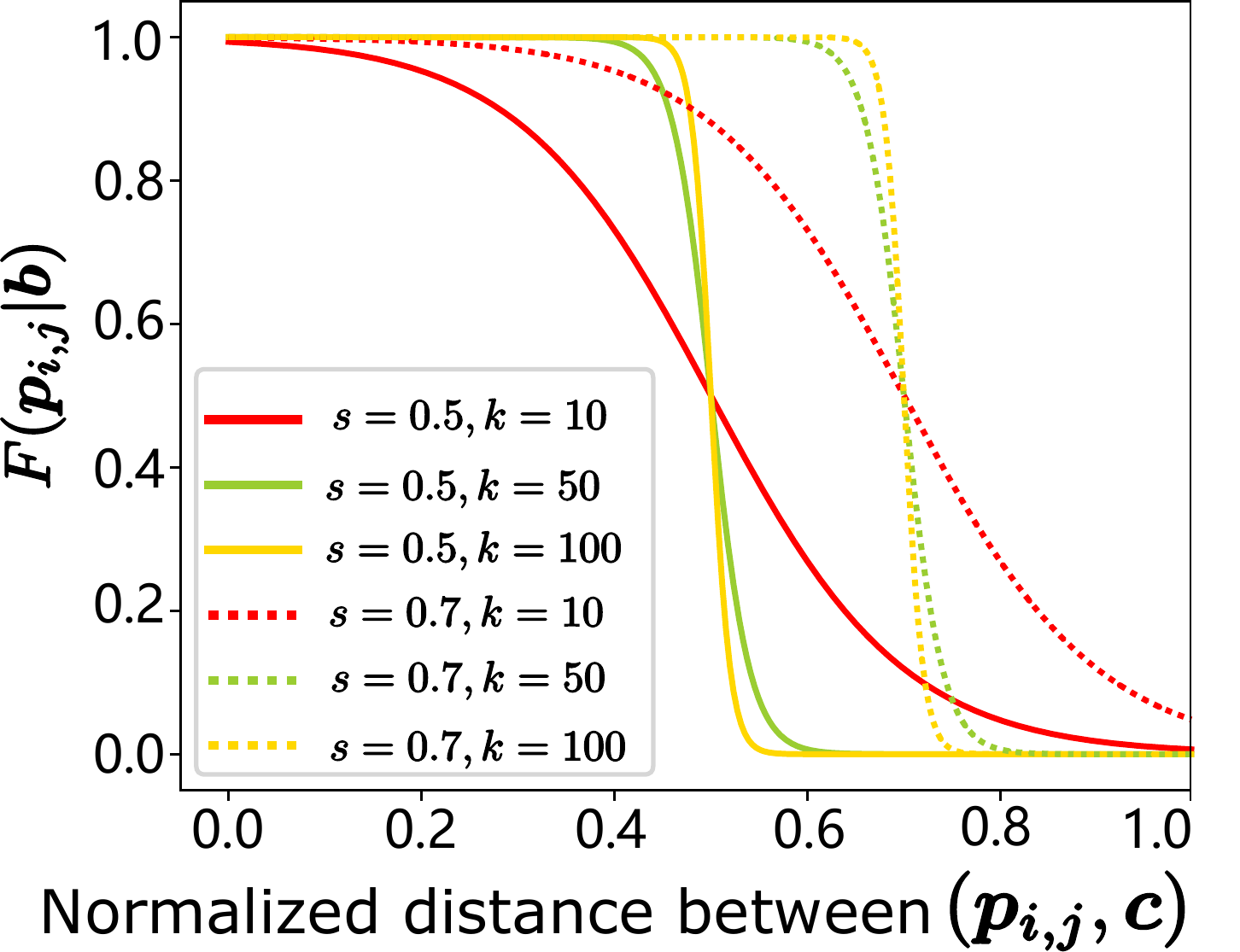}}
        \vspace{-3mm}
        \caption{General idea of the IoU function. (a) Components involved in determining the relative position (inside or outside) between a pixel $\bm{p}$ (green point) and an OBB $\bm{b}$ (red rectangle). Best viewed in color. (b) Distribution of the kernelized pixel contribution $F(\bm{p}_{i,j}|\bm{b})$ with different distances between $\bm{p}_{i,j}$ and box center $\bm{c}$. We see that $F(\bm{p}_{i,j}|\bm{b})$ is continuous and differentiable due to Eq.~\ref{equ:model:kernel}. Moreover, it approximately reflects the value distribution in Eq.~\ref{con:inside_pixel} when the pixels $\bm{p}_{i,j}$ are inside and outside $\bm{b}$.}
        \vspace{-3mm}
    \label{fig:model:piou}
\end{figure}

In this section, we present in detail the PIoU Loss.
For a given OBB $\bm{b}$ encoded by $(c_{x}, c_{y}, w, h, \theta)$, an ideal loss function should effectively guide the network to maximize the IoU and thereby the error of $\bm{b}$ can be minimized. Towards this goal, we first explain the IoU method. Generally speaking, an IoU function should accurately compute the area of an OBB as well as its intersection with another box. Since OBB and the intersection area are constructed by pixels in image space, their areas are approximated by the number of interior pixels. Specifically, as shown in Figure~\ref{fig:box_model}, $\bm{t}_{i,j}$ (the purple point) is the intersection point between the mid-vertical line and its perpendicular line to pixel $\bm{p}_{i,j}$ (the green point). As a result, a triangle is constructed by OBB center $\bm{c}$ (the red point), $\bm{p}_{i,j}$ and $\bm{t}_{i,j}$. The length of each triangle side is denoted by $d_{i,j}^w$, $d_{i,j}^h$ and $d_{i,j}$. To judge the relative location (inside or outside) between $\bm{p}_{i,j}$ and $\bm{b}$, we define the binary constraints as follows:
\begin{equation}
\delta(\bm{p}_{i,j}|\bm{b})=\left\{
\begin{aligned}
1, & \quad d_{i,j}^w \leq \frac{w}{2},d_{i,j}^h \leq \frac{h}{2} \\
0, & \quad otherwise
\end{aligned}
\right.
\label{con:inside_pixel}
\end{equation}
where $d_{ij}$ denotes the L2-norm distance between pixel~$(i, j)$ and OBB center~$(c_x, c_y)$, $d_w$ and $d_h$ denotes the distance $d$ along horizontal and vertical direction respectively:
\begin{align}
\small
d_{ij}= & \ d(i,j)=\sqrt{(c_x-i)^2+(c_y-j)^2} \\
d_{ij}^w= & \ |d_{ij}\cos\beta| \\
d_{ij}^h= & \ |d_{ij}\sin\beta| \\
\beta= & \left\{
\begin{aligned}
\theta+\arccos\frac{c_x-i}{d_{ij}}, & \quad c_y - j \geq 0 \\
\theta-\arccos\frac{c_x-i}{d_{ij}}, & \quad c_y - j \textless 0\\
\end{aligned}
\right.
\label{eq:distances}
\end{align}

Let \(B_{\bm{b},\bm{b}^{\prime}}\) denotes the smallest horizontal bounding box that covers both \(\bm{b}\) and \(\bm{b}^{\prime}\). We can then compute the intersection area $S_{\bm{b} \cap \bm{b}^{\prime}}$ and union area $S_{\bm{b} \cup \bm{b}^{\prime}}$ between two OBBs $\bm{b}$ and $\bm{b}^{\prime}$ using the statistics of all pixels in \(B_{\bm{b},\bm{b}^{\prime}}\):
\begin{align}
    S_{\bm{b} \cap \bm{b}^{\prime}} = \sum_{\bm{p}_{i,j}\in B_{\bm{b},\bm{b}^{\prime}}}&\delta(\bm{p}_{i,j}|\bm{b})\delta(\bm{p}_{i,j}|\bm{b}^{\prime}) \\
    S_{\bm{b} \cup \bm{b}^{\prime}}= \sum_{\bm{p}_{i,j}\in B_{\bm{b},\bm{b}^{\prime}}}\delta(\bm{p}_{i,j}|\bm{b})+&\delta(\bm{p}_{i,j}|\bm{b}^{\prime})-\delta(\bm{p}_{i,j}|\bm{b})\delta(\bm{p}_{i,j}|\bm{b}^{\prime})
\label{equ:model:oldious}
\end{align}

The final IoU of $\bm{b}$ and $\bm{b}^{\prime}$ can be calculated by dividing $S_{\bm{b} \cap \bm{b}^{\prime}}$ and $S_{\bm{b} \cup \bm{b}^{\prime}}$. However, we observe that Eq.~\ref{con:inside_pixel} is not a continuous and differentiable function. As a result, back propagation (BP) cannot utilize an IoU-based loss for training. To solve this problem, we approximate Eq.~\ref{con:inside_pixel} as $F(\bm{p}_{i,j}|\bm{b})$ taking on the product of two kernels:
\begin{equation}
\begin{split}
F(\bm{p}_{i,j}|\bm{b}) = K(d_{i,j}^w, w)K(d_{i,j}^h, h) \\
\end{split}
\label{equ:model:pixelgate}
\end{equation}
Particularly, the kernel function $K(d, s)$ is calculated by:
\begin{equation}
K(d,s)=1-\frac{1}{1+e^{-k(d-s)}}
\label{equ:model:kernel}
\end{equation}
where $k$ is an adjustable factor to control the sensitivity of the target pixel $\bm{p}_{i,j}$. The key idea of Eq.~\ref{equ:model:pixelgate} is to obtain the contribution of pixel $\bm{p}_{i,j}$ using the kernel function in Eq.~\ref{equ:model:kernel}. Since the employed kernel is calculated by the relative position (distance and angle of the triangle in Figure~\ref{fig:box_model}) between $\bm{p}_{i,j}$ and $\bm{b}$, the intersection area $S_{\bm{b} \cap \bm{b}^{\prime}}$ and union area $S_{\bm{b} \cup \bm{b}^{\prime}}$ are inherently sensitive to both OBB rotation and size. In Figure~\ref{fig:pixel_au}, we find that $F(\bm{p}_{i,j}|\bm{b})$ is continuous and differentiable. More importantly, it functions similarly to the characteristics of Eq.~\ref{con:inside_pixel} such that $F(\bm{p}_{i,j}|\bm{b})$ is close to 1.0 when the pixel $\bm{p}_{i,j}$ is inside and otherwise when $F(\bm{p}_{i,j}|\bm{b})\sim 0$. Following Eq.~\ref{equ:model:pixelgate}, the intersection area $S_{\bm{b} \cap \bm{b}^{\prime}}$ and union area $S_{\bm{b} \cup \bm{b}^{\prime}}$ between $\bm{b}$ and $\bm{b}^{\prime}$ are approximated by:
\begin{align}
\label{equ:model:newious}
    S_{\bm{b} \cap \bm{b}^{\prime}} \approx \sum_{\bm{p}_{i,j}\in B_{\bm{b},\bm{b}^{\prime}}}&F(\bm{p}_{i,j}|\bm{b})F(\bm{p}_{i,j}|\bm{b}^{\prime}) \\
    \label{equ:S_union}
    S_{\bm{b} \cup \bm{b}^{\prime}} \approx  \sum_{\bm{p}_{i,j}\in B_{\bm{b},\bm{b}^{\prime}}}F(\bm{p}_{i,j}|\bm{b})+&F(\bm{p}_{i,j}|\bm{b}^{\prime})-F(\bm{p}_{i,j}|\bm{b})F(\bm{p}_{i,j}|\bm{b}^{\prime})
\end{align}
In practice, to reduce the computational complexity of Eq.~\ref{equ:S_union}, $S_{\bm{b} \cup \bm{b}^{\prime}}$ can be approximated by a simpler form:
\begin{equation}
S_{\bm{b} \cup \bm{b}^{\prime}} = w \times h + w^{\prime} \times h^{\prime} - S_{\bm{b} \cap \bm{b}^{\prime}}
\label{equ:model:optmize}
\end{equation}
where $(w,h)$ and $(w^{\prime},h^{\prime})$ are the size of OBBs $\bm{b}$ and $\bm{b}^{\prime}$, respectively. Our experiment in Section~\ref{exp:evaluation:anchorbased} shows that Eq.~\ref{equ:model:optmize} can effectively reduce the complexity of Eq.~\ref{equ:model:newious} while preserving the overall detection performance. With these terms, our proposed Pixels-IoU $(PIoU)$ is computed as:
\begin{equation}
PIoU(\bm{b},\bm{b}^{\prime}) =\frac{S_{\bm{b} \cap \bm{b}^{\prime}}}{S_{\bm{b} \cup \bm{b}^{\prime}}}
\label{equ:model:piou}
\end{equation}

Let $\bm{b}$ denotes the predicted box and $\bm{b}^{\prime}$ denotes the ground-truth box. A pair $(\bm{b},\bm{b}^{\prime})$ is regarded as positive if the predicted box $\bm{b}$ is based on a positive anchor and \(\bm{b}^{\prime}\) is the matched ground-truth box (an anchor is matched with a ground-truth box if the IoU between them is larger them 0.5). We use \(M\) to denote the set of all positive pairs. With the goal to maximize the PIoU between $\bm{b}$ and $\bm{b}^{\prime}$, the proposed PIoU Loss is calculated by:
\begin{equation}
L_{piou}=\frac{-\sum_{(\bm{b},\bm{b}^{\prime})\in M}\ln{PIoU(\bm{b},\bm{b}^{\prime})}}{|M|}
\label{equ:model:finalloss}
\end{equation}

Theoretically, Eq.~\ref{equ:model:finalloss} still works if there is no intersection between $\bm{b}$ and $\bm{b}^{\prime}$. This is because $PIoU(\bm{b},\bm{b}^{\prime}) > 0$ based on Eq.~\ref{equ:model:kernel} and the gradients still exist in this case. Moreover, the proposed PIoU also works for horizontal bounding box regression. Specifically, we can simply set $\theta=0$ in Eq.~\ref{eq:distances} for this purpose. In Section~\ref{exp}, we experimentally validate the usability of PIoU for horizontal bounding box regression.

%% file: retail50k.tex
\section{Retail50K Dataset}
\label{exp:dataset}
\begin{figure}[t]
  \centering
    \includegraphics[width=0.7\linewidth]{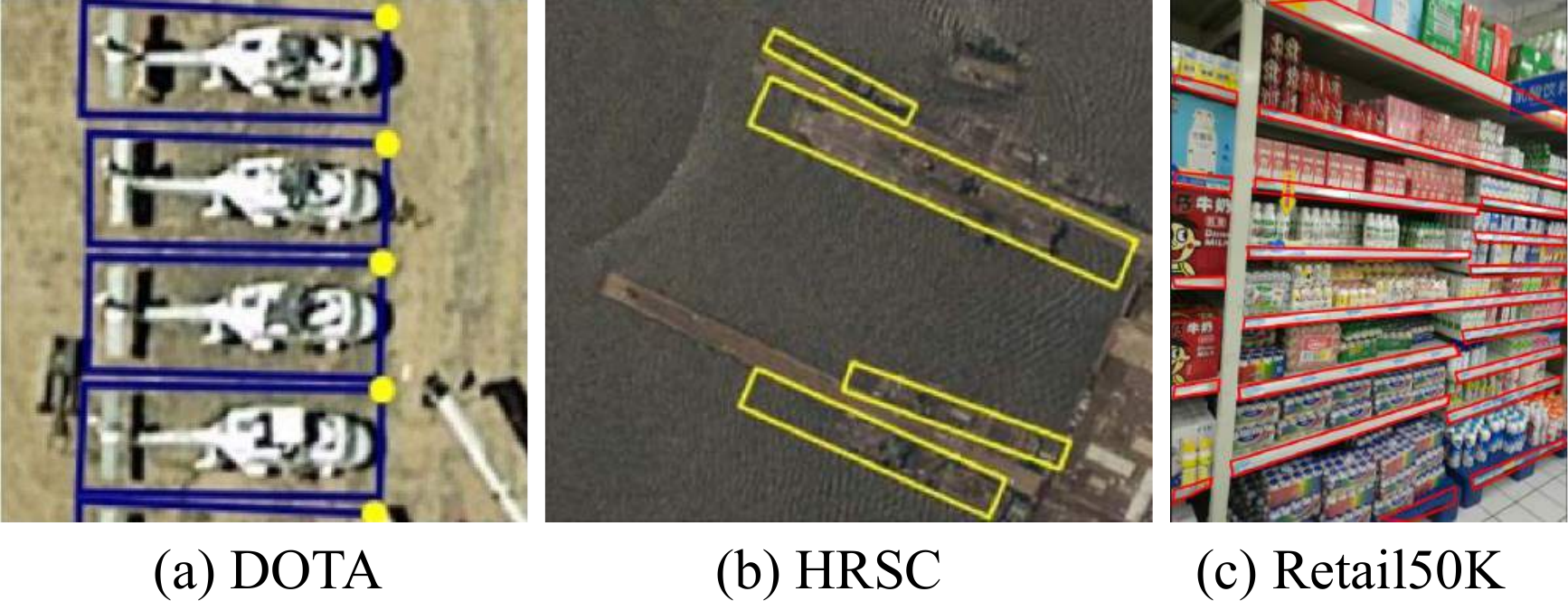}
  \vspace{-3mm}
  \caption{Sample images and their annotations of three datasets evaluated in our experiments: (a) DOTA~\cite{Xia2018DAS} (b) HRSC2016~\cite{Liu2016SRB} (c) Retail50K. There are two unique characteristics of Retail50K: (1) Complex backgrounds such as occlusions (by price tags), varied colours and textures. (2) OBB with high aspect ratios.}
  \vspace{-1mm}
\label{fig:exp:dataset_compare}
\end{figure}

OBB detectors have been actively studied for many years and several datasets with such annotations have been proposed~\cite{Xia2018DAS,Benedek2012BDM,Liu2015FMV,Liu2016SRB,Razakarivony2016VDI,Zhu2015ORO,Li2018MRB,He2015OOP}. As shown in Table~\ref{tab:multidatasetcompare}, most of them only focused on aerial images (Figure~\ref{fig:exp:dataset_compare} (a),(b)) while a few are annotated based on existing datasets such as MSCOCO~\cite{Lin2014MCC}, PASCAL VOC~\cite{Everingham2015TPV} and ImageNet~\cite{Deng2009IAH}. These datasets are important to evaluate the detection performance with simple backgrounds and low aspect ratios. For example, aerial images are typically gray and texture-less. The statistics in~\cite{Xia2018DAS} shows that most datasets of aerial images have a wide range of aspect ratios, but around 90\% of these ratios are distributed between 1:1 and 1:4, and very few images contain OBBs with aspect ratios larger than 1:5. Moreover, aspect ratios of OBBs on PASCAL VOC are mostly close to square (1:1). As a result, it is hard to assess the capability of detectors on objects with high aspect ratios and complex backgrounds using existing datasets. Motivated by this, we introduce a new dataset, namely Retail50K, to advance the research of detection of rotated objects in complex environments. We intend to make this publicly available to the community (\url{https://github.com/clobotics/piou}).
\begin{table}[t]
\small
\centering
\setlength{\tabcolsep}{5pt}
\caption{Comparison between different datasets with OBB annotations. $\approx$ indicate estimates based on selected annotated samples as full access was not possible.}
\vspace{-2mm}
\begin{tabular}{l|ccccc}
\hline
Dataset                                  & Scenario   & Median Ratio & Images & Instances \\ \hline
SZTAKI~\cite{Benedek2012BDM}             & Aerial     & $\approx$1:3     & 9      & 665       \\
VEDAI~\cite{Razakarivony2016VDI}         & Aerial     & 1:3     & 1268   & 2950      \\
UCAS-AOD~\cite{Zhu2015ORO}               & Aerial     & 1:1.3   & 1510   & 14596     \\
HRSC2016~\cite{Liu2016SRB}               & Aerial     & 1:5     & 1061   & 2976      \\
Vehicle~\cite{Liu2015FMV}                & Aerial     & 1:2     & 20     & 14235     \\
DOTA~\cite{Xia2018DAS}                   & Aerial     & 1:2.5     & 2806   & 188282    \\
SHIP~\cite{Li2018MRB}                    & Aerial     & $\approx$1:5     & 640    & -    \\
OOP~\cite{He2015OOP}                     & PASCAL     & $\approx$1:1     & 4952   & -    \\
\textbf{Proposed}                        & \textbf{Retail}  & \textbf{1:20} & \textbf{47000}  & \textbf{48000}  \\ \hline
\end{tabular}
\label{tab:multidatasetcompare}
\end{table}

\begin{figure*}[t]
  \centering
    \includegraphics[width=1\linewidth]{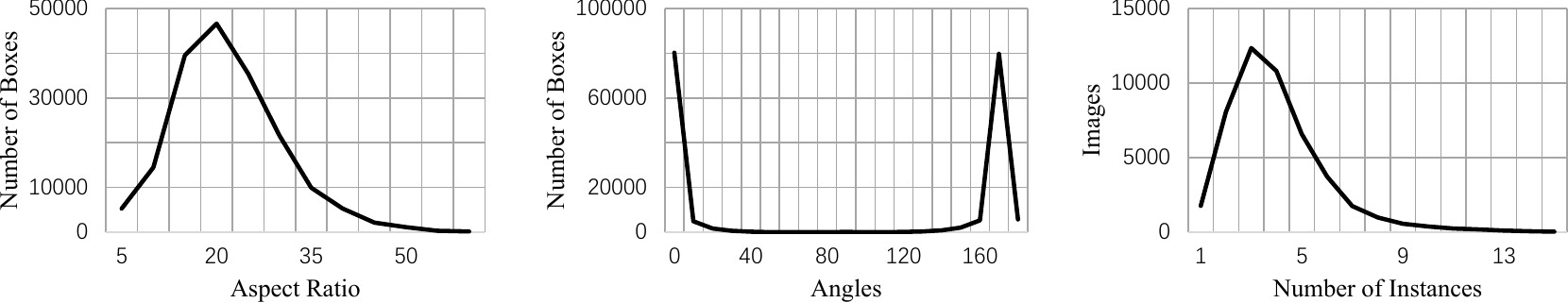}
  \vspace{-3mm}
  \caption{Statistics of different properties of Retail50K dataset.}
  \vspace{-3mm}
\label{fig:exp:statistics}
\end{figure*}

Figure~\ref{fig:exp:dataset_compare} (c) illustrates a sample image from Retail50K dataset. Retail50K is a collection of 47,000 images from different supermarkets. Annotations on those images are the layer edges of shelves, fridges and displays. We focus on such retail environments for three reasons: (1) \textbf{Complex background.} Shelves and fridges are tightly filled with many different items 
with a wide variety of colours and textures. Moreover, layer edges are normally occluded by price tags and sale tags. Based on our statistics, the mean occlusion is around 37.5\%. It is even more challenging that the appearance of price tags are different in different supermarkets. (2) \textbf{High aspect ratio.} Aspect ratio is one of the essential factors for anchor-based models~\cite{Redmon2016YBF}. Bounding boxes in Retail50K dataset not only have large variety in degrees of orientation, but also a wide range of aspect ratios. In particular, the majority of annotations in Retail50K are with high aspect ratios. Therefore, this dataset represents a good combination of challenges that is precisely the type we find in complex retail environments.
(3) \textbf{Useful in practice.} The trained model based on Retail50K can be used for many applications in retail scenarios such as shelf retail tag detection, automatic shelf demarcation, shelf layer and image yaw angle estimation, etc. It is worth to note that although SKU-110K dataset~\cite{Goldman2019PDI} is also assembled from retail environment such as supermarket shelves, the annotations in this dataset are horizontal bounding boxes (HBB) of shelf products since it mainly focuses on object detection in densely packed scenes. The aspect ratios of its HBB are distributed between 1:1-1:3 and hence, it does not cater to the problem that we want to solve. 

\noindent{\bf Images and Categories:} Images in Retail50K were collected from 20 supermarket stores in China and USA. Dozens of volunteers acquired data using their personal cellphone cameras. To increase the diversity of data, images were collected in multiple cities from different volunteers. Image quality and view settings were unregulated and so the collected images represent different scales, viewing angles, lighting conditions, noise levels, and other sources of variability. We also recorded the meta data of the original images such as capture time, volunteer name, shop name and MD5~\cite{Xie2013FCA} checksum to filter out duplicated images. Unlike existing datasets that contain multiple categories~\cite{Xia2018DAS,Lin2014MCC,Everingham2015TPV,Deng2009IAH}, there is only one category in Retail50K dataset. For better comparisons across datasets, we also employ DOTA~\cite{Xia2018DAS} (15 categories) and HRSC2016~\cite{Liu2016SRB} (the aspect ratio of objects is between that of Retail50K and DOTA) in our experiments (Figure~\ref{fig:exp:dataset_compare}).

\noindent{\bf Annotation and Properties:} In Retail50K dataset, bounding box annotations were provided by 5 skilled annotators. To improve their efficiency, a handbook of labelling rules was provided during the training process. Candidate images were grouped into 165 labelling tasks based on their meta-data so that peer reviews can be applied. Finally, considering the complicated background and various orientations of layer edges, we perform the annotations using arbitrary quadrilateral bounding boxes (AQBB). Briefly, AQBB is denoted by the vertices of the bounding polygon in clockwise order. Due to high efficiency and empirical success, AQBB is widely used in many benchmarks such as text detection~\cite{Karatzas2015COR}, object detection in aerial images~\cite{Li2018MRB}, etc. Based on AQBB, we can easily compute the required OBB format which is denoted by $(c_{x},c_{y},w,h,\theta)$.

Since images were collected with personal cellphone cameras, the original images have different resolutions; hence they were uniformly resized into $600 \times 800$ before annotation took place. Figure~\ref{fig:exp:statistics} shows some statistics of Retail50K. We see that the dataset contains a wide range of aspect ratios and orientations (Figure~\ref{fig:exp:statistics} (a) and (b)). In particular, Retail50K is more challenging as compared to existing datasets~\cite{Liu2015FMV,Xia2018DAS,Li2018MRB} since it contains rich annotations with extremely high aspect ratios (higher than 1:10). Similar to natural-image datasets such as ImageNet (average 2) and MSCOCO (average 7.7), most images in our dataset contain around 2-6 instances with complex backgrounds (Figure~\ref{fig:exp:statistics} (c)). For experiments, we selected half of the original images as the training set, 1/6 as validation set, and 1/3 as the testing set.

%% file: experiment.tex
\section{Experiments}
\label{exp}
\subsection{Experimental Settings}
We evaluate the proposed PIoU loss with anchor-based and anchor-free OBB-detectors (RefineDet, CenterNet) under different parameters, backbones. We also compare the proposed method with other state-of-the-art OBB-detection methods in different benchmark datasets (\textit{i.e.} DOTA~\cite{Xia2018DAS}, HRSC2016~\cite{Liu2016SRB}, PASCAL VOC~\cite{Everingham2015TPV}) and the proposed Retail50K dataset. The training and testing tasks are accomplished on a desktop machine with Intel(R) Core(TM) i7-6850K CPU @ 3.60GHzs, 64 GB installed memory, a GeForce GTX 1080TI GPU (11 GB global memory), and Ubuntu 16.04 LTS. With this machine, the batch size is set to 8 and 1 for training and testing, respectively.

\noindent\textbf{Anchor-based OBB Detector:}
\label{approach:anchorbased}
For anchor-based object detection, we train RefineDet~\cite{refinedet} by updating its loss using the proposed PIoU method. Since the detector is optimized by classification and regression losses, we can easily replace the regression one with PIoU loss $L_{piou}$ while keeping the original Softmax Loss $L_{cls}$ for classification. We use ResNet~\cite{He2016DRL} and VGG ~\cite{Simonyan2014VDC} as the backbone models. The oriented anchors are generated by rotating the horizontal anchors by \(k\pi/6\) for \(0\leq k<6\). We adopt the data augmentation strategies introduced in \cite{Liu2016SSS} except cropping, while including rotation (i.e. rotate the image by a random angle sampled in \([0,\pi/6]\)). In training phase, the input image is resized to 512\(\times\)512. We adopt the mini-batch training on 2 GPUs with 8 images per GPU. SGD is adopted to optimize the models with momentum set to 0.9 and weight decay set to 0.0005. All evaluated models are trained for 120 epochs with an initial learning rate of 0.001 which is then divided by 10 at 60 epochs and again at 90 epochs. Other experimental settings are the same as those in \cite{refinedet}.

\noindent\textbf{Anchor-free OBB Detector:}
\label{approach:anchorfree}
To extend anchor-free frameworks for detecting OBB, we modify CenterNet~\cite{Zhou2019OAP} by adding an angle dimension regressed by L1-Loss in its overall training objective as our baseline. To evaluate the proposed loss function, in similar fashion as anchor-based approach, we can replace the regression one with PIoU loss $L_{piou}$ while keeping the other classification loss $L_{cls}$ the same. Be noted that CenterNet uses a heatmap to locate the center of objects. Thus, we do not back-propagate the gradient of the object's center when computing the PIoU loss. We use DLA~\cite{Yu2018Deep} and ResNet~\cite{He2016DRL} as the backbone models. The data augmentation strategies is the same as those for RefineDet-OBB (shown before). In training phase, the input image is resized to 512\(\times\)512. We adopt the mini-batch training on 2 GPUs with 16 images per GPU. ADAM is adopted to optimize the models. All evaluated models are trained for 120 epochs with an initial learning rate of 0.0005 which is then divided by 10 at 60 epochs and again at 90 epochs. Other settings are the same as those in \cite{Zhou2019OAP}.

\subsection{Ablation Study}
\label{exp:evaluation:anchorbased}
In this section, we investigate the impact of our design settings of the proposed method, and conduct several controlled experiments on DOTA~\cite{Xia2018DAS} and PASCAL VOC~\cite{Everingham2015TPV} datasets.

\noindent{\bf Comparison on different parameters:}
In Eq.~\ref{equ:model:kernel}, $k$ is an adjustable factor in our kernel function to control the sensitivity of each pixel. In order to evaluate its influence as well as to find a proper value for the remaining experiments, we conduct a set of experiments by varying $k$ values based on DOTA~\cite{Xia2018DAS} dataset with the proposed anchor-based framework. To simplify discussions, results of $k=5, 10, 15$ are detailed in Table~\ref{tab:Different_hyperparameter} while their distributions can be visualized in Fig.~\ref{fig:pixel_au}. We finally select $k=10$ for the rest of the experiments since it achieves the best accuracy.

\noindent{\bf Comparison for oriented bounding box:} Based on DOTA~\cite{Xia2018DAS} dataset, we compare the proposed PIoU loss with the commonly used L1 loss, SmoothL1 loss as well as L2 loss. For fair comparisons, we fix the backbone to VGGNet~\cite{Simonyan2014VDC} and build the network based on FPN~\cite{Lin2017FPN}. Table~\ref{tab:losses:dota} details the comparisons and we can clearly see that the proposed PIoU Loss improves the detection performance by around 3.5\%. HPIoU (Hard PIoU) loss is the simplified PIoU loss using Eq.~\ref{equ:model:optmize}. Its performance is slightly reduced but still comparable to PIoU loss. Thus, HPIoU loss can be a viable option in practise as it has lower computational complexity. We also observe that the proposed PIoU costs 15-20\% more time than other three loss functions, which shows that it is still acceptable in practice. We also observed that HPIoU costs less training time than PIoU. Such observation verifies the theoretical analysis and usability of Eq.~\ref{equ:model:optmize}.

\begin{table}[t!]
\small
\centering
\setlength{\tabcolsep}{20pt}
\caption{Comparison between different sensitivity factor $k$ in Eq.~\ref{equ:model:kernel} for PIoU loss on DOTA dataset. RefineDet~\cite{refinedet} is used as the detection model.}
\vspace{-1mm}
\begin{tabular}{cccc}
\hline
$k$    & AP   & AP\(_{50}\) & AP\(_{75}\)  \\ \hline
5  & 46.88 & 59.03 &  34.73 \\
10 & 54.24 & 67.89 & 40.59  \\
15  & 53.41& 65.97 & 40.84  \\ \hline
\end{tabular}
\vspace{-0.2em}
\label{tab:Different_hyperparameter}
\end{table}
\begin{table}[t!]
\small
\centering
\setlength{\tabcolsep}{12pt}
\caption{Comparison between different losses for oriented bounding box on DOTA dataset. RefineDet~\cite{refinedet} is used as the detection model. HPIoU (Hard PIoU) loss refers to the PIoU loss simplified by Eq.~\ref{equ:model:optmize}. Training time is estimated in hours.}
\vspace{-1mm}
\begin{tabular}{lcccc}
\hline
Loss    & AP   & AP\(_{50}\) & AP\(_{75}\) & Training Time  \\ \hline
L1 Loss     & 50.66 & 64.14 &  37.18 & 20 \\
L2 Loss     & 49.70 & 62.74 &  36.65 & 20 \\
SmoothL1 Loss & 51.46 & 65.68 & 37.25 & 21.5 \\
\textbf{PIoU Loss} & \textbf{54.24} & \textbf{67.89} & \textbf{40.59} & \textbf{25.7}  \\
\textbf{HPIoU Loss} & \textbf{53.37} & \textbf{66.38} & \textbf{40.36} & \textbf{24.8} \\ \hline
\end{tabular}
\vspace{-0.2em}
\label{tab:losses:dota}
\end{table}
\begin{table}[t!]
\small
\centering
\setlength{\tabcolsep}{7pt}
\caption{Comparison between different losses for horizontal bounding box on PASCAL VOC2007 dataset. SSD~\cite{Liu2016SSS} is used as the detection model.}
\vspace{-1mm}
\begin{tabular}{lccccccccccc}
\hline
Loss    & AP   & AP\(_{50}\) & AP\(_{60}\) & AP\(_{70}\) &AP\(_{80}\) & AP\(_{90}\) \\
\hline
SmoothL1 Loss & 48.8 & 79.8 & 72.9 & 60.6 & 40.3 & 10.2    \\
GIoU Loss~\cite{Rezatofighi2019GIO} & 49.9 & 79.8 & 74.1 & 63.2 & 41.9 & 12.4 \\
\textbf{PIoU Loss} & \textbf{50.3} & \textbf{80.1} & \textbf{74.9} & \textbf{63.0} & \textbf{42.5} & \textbf{12.2}  \\
\hline
\end{tabular}
\vspace{-3mm}
\label{tab:losses:voc}
\end{table}

\noindent{\bf Comparison for horizontal bounding box:} Besides, we also compare the PIoU loss with SmoothL1 loss and GIoU loss~\cite{Rezatofighi2019GIO} for horizontal bounding box on PASCAL VOC dataset~\cite{Everingham2015TPV}. In Table \ref{tab:losses:voc}, we observe that the proposed PIoU loss is still better than SmoothL1 loss and GIoU loss for horizontal bounding box regression, particularly at those AP metrics with high IoU threshold. Note that the GIoU loss is designed only for horizontal bounding box while the proposed PIoU loss is more robust and well suited for both horizontal and oriented bounding box. Together with the results in Table \ref{tab:losses:dota}, we observe the strong generalization ability and effectiveness of the proposed PIoU loss.

\subsection{Benchmark Results}
\begin{table}[t]
\footnotesize
\centering
\setlength{\tabcolsep}{3pt}
\caption{Detection results on Retail50K dataset. The PIoU loss is evaluated on RefineDet~\cite{refinedet} and CenterNet~\cite{Zhou2019OAP} with different backbone models.}
\vspace{-2mm}
\begin{tabular}{lcccccc}
\hline
Method               & Backbone    & AP    & AP\(_{50}\) & AP\(_{75}\) & Time (ms) & FPS  \\ \hline
RefineDet-OBB~\cite{refinedet}     & ResNet-50 & 53.96 & 74.15 & 33.77 & 142 & 7  \\
\textbf{RefineDet-OBB+PIoU} & ResNet-50 & \textbf{61.78} & \textbf{80.17} & \textbf{43.39} & \textbf{142} & \textbf{7}   \\
RefineDet-OBB~\cite{refinedet} & ResNet-101      & 55.46 & 77.05 & 33.87 & 167 & 6   \\
\textbf{RefineDet-OBB+PIoU} & ResNet-101 & \textbf{63.00} & \textbf{79.08} & \textbf{46.01} & \textbf{167} & \textbf{6}   \\ \hline
CenterNet-OBB~\cite{Zhou2019OAP} & ResNet18      & 54.44 & 76.58  & 32.29  & 7 & 140 \\
\textbf{CenterNet-OBB+PIoU} & ResNet18 & \textbf{61.02} & \textbf{87.19}  & \textbf{34.85}  & \textbf{7} & \textbf{140} \\
CenterNet-OBB~\cite{Zhou2019OAP} & DLA-34      & 56.13 & 78.29  & 33.97  & 18.18 & 55 \\
\textbf{CenterNet-OBB+PIoU} & DLA-34 & \textbf{61.64} & \textbf{88.47}  & \textbf{34.80}  & \textbf{18.18} & \textbf{55} \\\hline
\end{tabular}
\label{tab:exp:eval:potential}
\end{table}
\begin{table}[t!]
\footnotesize
\centering
\setlength{\tabcolsep}{5pt}
\caption{Detection results on HRSC2016 dataset. \emph{Aug.} indicates data augmentation. \emph{Size} means the image size that used for training and testing.}
\vspace{-2mm}
\begin{tabular}{llcccc}
\hline
Method                 & Backbone    & Size & Aug. & mAP &FPS  \\ \hline
R$^2$CNN~\cite{Jiang2017RRR}               & ResNet101   & 800 $\times$ 800    & $\times$         & 73.03 & 2\\
RC1 \& RC2~\cite{liu2017high} & VGG-16 & - & - & 75.7 & \(<\)1fps \\
RRPN~\cite{Ma2018AST}                   & ResNet101   & 800 $\times$ 800    & $\times$         & 79.08 & 3.5\\
R\(^2\)PN~\cite{zhang2018toward} & VGG-16 & - & \(\surd\) & 79.6 & \(<\)1fps \\
RetinaNet-H~\cite{Yang2019RRS}            & ResNet101   & 800 $\times$ 800    & $\surd$         & 82.89 & 14\\
RetinaNet-R~\cite{Yang2019RRS}            & ResNet101   & 800 $\times$ 800    & $\surd$         & 89.18 & 10\\
RoI-Transformer~\cite{Jian2019LRT}        & ResNet101   & 512 $\times$ 800    & $\times$         & 86.20 & - \\ \hline
\multirow{3}{*}{R$^3$Det~\cite{Yang2019RRS}}
        & ResNet101   & 300 $\times$ 300    & $\surd$  & 87.14 & 18\\
        & ResNet101   & 600 $\times$ 600    & $\surd$  & 88.97 & 15\\
        & ResNet101   & 800 \(\times\) 800  & \(\surd\) & 89.26 & 12\\
        \hline \hline
CenterNet-OBB~\cite{Zhou2019OAP}       & ResNet18   & 512 $\times$ 512    & $\surd$         & 67.73 & 140\\
\textbf{CenterNet-OBB+PIoU}   & \textbf{ResNet18}   & \textbf{512 $\times$ 512}    & $\surd$         & \textbf{78.54} & \textbf{140}\\
CenterNet-OBB~\cite{Zhou2019OAP}       & ResNet101   & 512 $\times$ 512    & $\surd$         & 77.43 & 45\\
\textbf{CenterNet-OBB+PIoU}   & \textbf{ResNet101}   & \textbf{512 $\times$ 512}    & $\surd$         & \textbf{80.32} & \textbf{45}\\
CenterNet-OBB~\cite{Zhou2019OAP}       & DLA-34   & 512 $\times$ 512    & $\surd$         & 87.98 & 55\\
\textbf{CenterNet-OBB+PIoU}             & \textbf{DLA-34}   & \textbf{512 $\times$ 512}    & $\surd$         & \textbf{89.20} & \textbf{55}\\ \hline
\end{tabular}
\vspace{-1mm}
\label{tab:state-of-the-art:hrsc2016}
\end{table}

\noindent\textbf{Retail50K:}
We evaluate our PIoU loss with two OBB-detectors (\textit{i.e.} the OBB versions of RefineDet~\cite{refinedet} and CenterNet~\cite{Zhou2019OAP}) on Retail50K dataset. The experimental results are shown in Table \ref{tab:exp:eval:potential}. We observe that, both detectors achieve significant improvements with the proposed PIoU loss (\(\sim\) 7\% improvement for RefineDet-OBB and \(\sim\) 6\% improvement for CenterNet-OBB). One reason for obtaining such notable improvements is that the proposed PIoU loss is much better suited for oriented objects than the traditional regression loss. Moreover, the improvements from PIoU loss in Retail50K are more obvious than those in DOTA (\textit{c.f.} Table \ref{tab:losses:dota}), which could mean that the proposed PIoU loss is extremely useful for objects with high aspect ratios and complex environments. This verifies the effectiveness of the proposed method.

\noindent\textbf{HRSC2016:}
The HRSC2016 dataset~\cite{Liu2016SRB} contains 1070 images from two scenarios including ships on sea and ships close inshore. We evaluate the proposed PIoU with CenterNet~\cite{Zhou2019OAP} on different backbones, and compare them with several state-of-the-art detectors. The experimental results are shown in Table \ref{tab:state-of-the-art:hrsc2016}. It can be seen that the CenterNet-OBB+PIoU outperforms all other methods except R\(^3\)Det-800. This is because we use a smaller image size (512\(\times\)512) than R\(^3\)Det-800 (800\(\times\)800). Thus, our detector preserves a reasonably competitive detection performance, but with far better efficiency (55 fps \textit{v.s} 12 fps). This exemplifies the strength of the proposed PIoU loss on OBB detectors.

\noindent\textbf{DOTA:}
The DOTA dataset~\cite{Xia2018DAS} contains 2806 aerial images from different sensors and platforms with crowd-sourcing. Each image is of size about 4000\(\times\)4000 pixels and contains objects of different scales, orientations and shapes. Note that image in DOTA is too large to be directly sent to CNN-based detectors. Thus, similar to the strategy in \cite{Xia2018DAS}, we crop a series of 512\(\times\)512 patches from the original image with the stride set to 256. For testing, the detection results are obtained from the DOTA evaluation server. The detailed performances for each category are reported so that deeper observations could be made. We use the same short names, benchmarks and forms as those existing methods in~\cite{Yang2019RRS} to evaluate the effectiveness of PIoU loss on this dataset. The final results are shown in Table~\ref{tab:state-of-the-art:dota}.
We find that the performance improvements vary among different categories. However, it is interesting to find that the improvement is more plausible for some categories with high aspect ratios. For example, harbour (HA), ground track field (GTF), soccer-ball field (SBF) and basketball court (BC) all naturally have large aspect ratios, and they appear to benefit from the inclusion of PIoU. Such observations confirm that the PIoU can effectively improve the performance of OBB detectors, particularly on objects with high-aspect ratios. These verify again the effectiveness of the proposed PIoU loss on OBB detectors. We also find that our baselines are relatively low than some state-of-the-art performances. We conjecture the main reason is that we use much smaller input size than other methods (512 vs 1024 on DOTA). However, note that the existing result (89.2 mAP) for HRSC2016 in Table~\ref{tab:state-of-the-art:hrsc2016} already achieves the state-of-the-art level performance with only $512\times512$ image size. Thus, the proposed loss function can bring gain in this strong baseline.

\begin{table*}[t!]
\tiny
\centering
\setlength{\tabcolsep}{0.7pt}
\caption{Detection results on DOTA dataset. We report the detection results for each category to better demonstrate where the performance gains come from.}
\vspace{-1mm}
\begin{tabular}{l|c|c|c|c|c|c|c|c|c|c|c|c|c|c|c|c|c|c}
\hline
Method & Backbone & Size & PL    & BD    & BR    & GTF   & SV    & LV    & SH    & TC    & BC    & ST    & SBF   & RA    & HA    & SP    & HC  &mAP   \\ \hline
SSD~\cite{Liu2016SSS}      & VGG16 & 512      & 39.8 & 9.1  & 0.6  & 13.2 & 0.3  & 0.4  & 1.1  & 16.2 & 27.6 & 9.2  & 27.2 & 9.1  & 3.0  & 1.1  & 1.0  & 10.6 \\
YOLOV2~\cite{Redmon2016YBF} & DarkNet19 & 416 & 39.6 & 20.3 & 36.6 & 23.4 & 8.9  & 2.1  & 4.8  & 44.3 & 38.4 & 34.7 & 16.0 & 37.6 & 47.2 & 25.5 & 7.5  & 21.4 \\
R-FCN~\cite{Dai2016ROD} & ResNet101 & 800 & 37.8 & 38.2 & 3.6 & 37.3 & 6.7 & 2.6 & 5.6 & 22.9 & 46.9 & 66.0 & 33.4 & 47.2 & 10.6 & 25.2 & 18.0 & 26.8 \\
FR-H~\cite{Ren2015FRT} & ResNet101 & 800 & 47.2 & 61.0 & 9.8 & 51.7 & 14.9 & 12.8 & 6.9 & 56.3 & 60.0 & 57.3 & 47.8 & 48.7 & 8.2 & 37.3 & 23.1 & 32.3 \\
FR-O~\cite{Xia2018DAS} & ResNet101 & 800 & 79.1 & 69.1 & 17.2 & 63.5 & 34.2 & 37.2 & 36.2 & 89.2 & 69.6 & 59.0 & 49. & 52.5 & 46.7 & 44.8 & 46.3 & 52.9 \\
R-DFPN~\cite{yang2018automatic} & ResNet101 & 800 & 80.9 & 65.8 & 33.8 & 58.9 & 55.8 & 50.9 & 54.8 & 90.3 & 66.3 & 68.7 & 48.7 & 51.8 & 55.1 & 51.3 & 35.9 & 57.9 \\
R\(^2\)CNN~\cite{Jiang2017RRR} & ResNet101 & 800 & 80.9 & 65.7 & 35.3 & 67.4 & 59.9 & 50.9 & 55.8 & 90.7 & 66.9 & 72.4 & 55.1 & 52.2 & 55.1 & 53.4 & 48.2 & 60.7 \\
RRPN~\cite{Ma2018AST} & ResNet101 & 800 & 88.5 & 71.2 & 31.7 & 59.3 & 51.9 & 56.2 & 57.3 & 90.8 & 72.8 & 67.4 & 56.7 & 52.8 & 53.1 & 51.9 & 53.6 & 61.0 \\
\hline \hline
RefineDet~\cite{refinedet} & VGG16  & 512 & 80.5 & 26.3 & 33.2 & 28.5 & 63.5 & 75.1 & 78.8 & 90.8 & 61.1 & 65.9 & 12.1 & 23.0 & 50.9 & 50.9 & 22.6 &  50.9 \\
\textbf{RefineDet+PIoU}  & VGG16 & 512 & \textbf{80.5} & \textbf{33.3} & \textbf{34.9} & \textbf{28.1} & \textbf{64.9} & \textbf{74.3} & \textbf{78.7} & \textbf{90.9} & \textbf{65.8} & \textbf{66.6} & \textbf{19.5} & \textbf{24.6} & \textbf{51.1} & \textbf{50.8} & \textbf{23.6}& \textbf{52.5} \\
RefineDet~\cite{refinedet}& ResNet101 & 512 & 80.7 & 44.2 & 27.5 & 32.8 & 61.2 & 76.1 & 78.8 & 90.7 & 69.9 & 73.9 & 24.9 & 31.9 & 55.8 & 51.4 & 26.8 & 55.1 \\
\textbf{RefineDet+PIoU} & ResNet101 & 512 & \textbf{80.7} & \textbf{48.8} & \textbf{26.1} & \textbf{38.7} & \textbf{65.2} & \textbf{75.5} & \textbf{78.6} & \textbf{90.8} & \textbf{70.4} & \textbf{75.0} & \textbf{32.0} & \textbf{28.0} & \textbf{54.3} & \textbf{53.7} & \textbf{29.6} & \textbf{56.5}\\
\hline
CenterNet~\cite{Zhou2019OAP} & DLA-34 & 512 & 81.0 & 64.0 & 22.6 & 56.6 & 38.6 & 64.0 & 64.9 & 90.8 & 78.0 & 72.5 & 44.0 & 41.1 & 55.5 & 55.0 & 57.4 & 59.1 \\
\textbf{CenterNet+PIoU} & DLA-34 & 512 & \textbf{80.9} & \textbf{69.7} & \textbf{24.1} & \textbf{60.2} & \textbf{38.3} & \textbf{64.4} & \textbf{64.8} & \textbf{90.9} & \textbf{77.2} & \textbf{70.4} & \textbf{46.5} & \textbf{37.1}  & \textbf{57.1} & \textbf{61.9} & \textbf{64.0} & \textbf{60.5} \\
\hline
\end{tabular}
\label{tab:state-of-the-art:dota}
\vspace{-2mm}
\end{table*}
\begin{figure*}[t]
  \centering
    \includegraphics[width=1\linewidth]{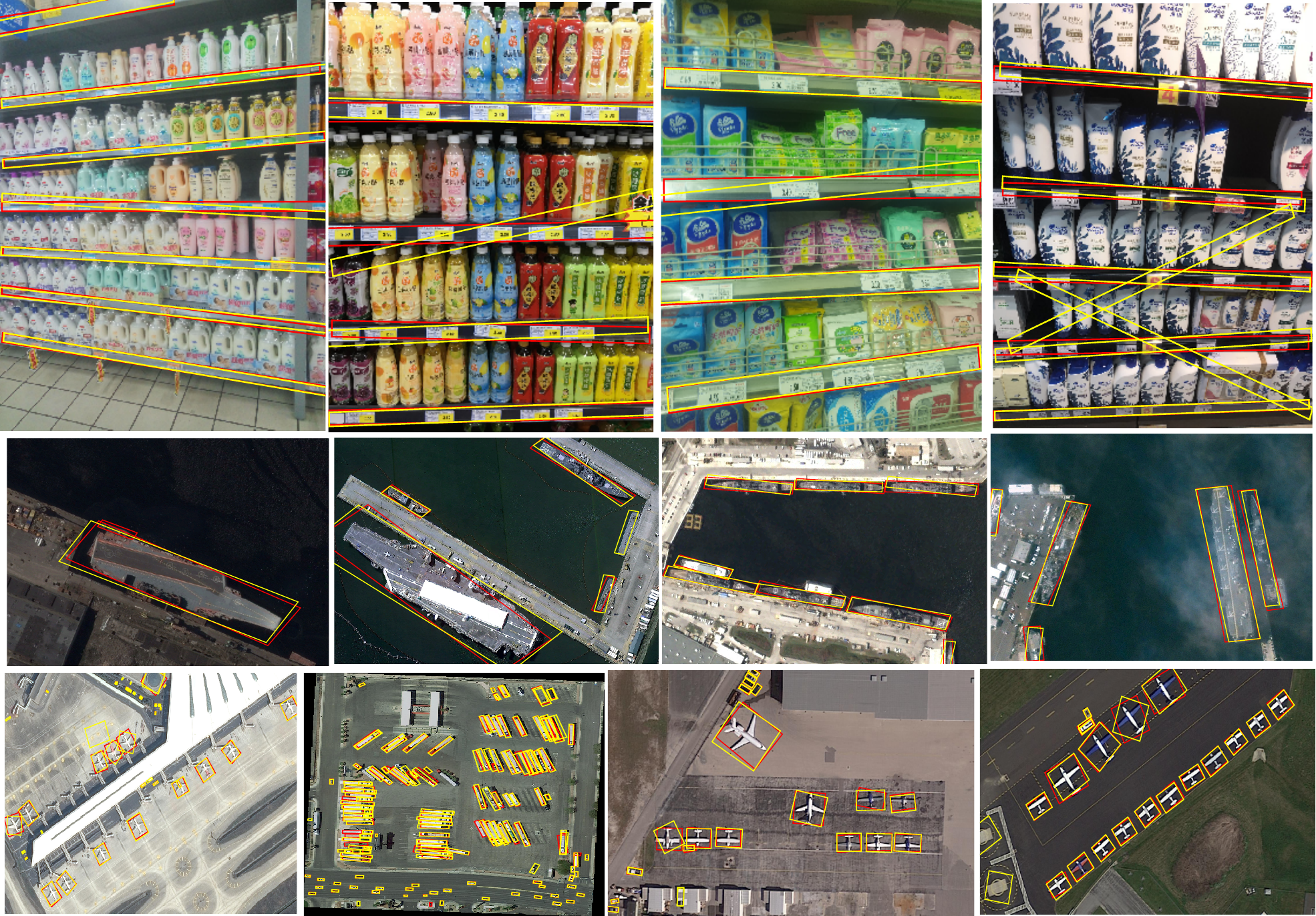}
  \vspace{-2mm}
  \caption{Samples results using PIoU (red boxes) and SmoothL1 (yellow boxes) losses on Retail50K (first row), HRSC2016 (second row) and DOTA (last row) datasets.}
\label{fig:exp:baseline:ourmethod:vis}
\vspace{-4mm}
\end{figure*}

In order to visually verify these performance improvements, we employ the anchor-based model RefineDet~\cite{refinedet} and conduct two independent experiments using PIoU and SmoothL1 losses. The experiments are applied on all three datasets (\textit{i.e.} Retail50K, DOTA~\cite{Xia2018DAS}, HRSC2016~\cite{Liu2016SRB}) and selected visual results are presented in Figure~\ref{fig:exp:baseline:ourmethod:vis}. We can observe that the OBB detector with PIoU loss (in red boxes) has more robust and accurate detection results than the one with SmoothL1 loss (in yellow boxes) on all three datasets, particularly on Retail50K, which demonstrates its strength in improving the performance for high aspect ratio oriented objects. Here, we also evaluate the proposed HPIoU loss with the same configuration of PIoU. In our experiments, the performances of HPIoU loss are slightly lower than those of PIoU loss (0.87, 1.41 and 0.18 mAP on DOTA, Retail50K and HRSC2016 respectively), but still better than smooth-L1 loss while having higher training speed than PIoU loss. Overall, the performances of HPIoU are consistent on all three datasets.

%% file: conclusion.tex
\section{Conclusion}
\label{sec:con}
We introduce a simple but effective loss function, PIoU, to exploit both the angle and IoU for accurate OBB regression. The PIoU loss is derived from IoU metric with a pixel-wise form, which is simple and suitable for both horizontal and oriented bounding box. To demonstrate its effectiveness, we evaluate the PIoU loss on both anchor-based and anchor-free frameworks. The experimental results show that PIoU loss can significantly improve the accuracy of OBB detectors, particularly on objects with high-aspect ratios. We also introduce a new challenging dataset, Retail50K, to explore the limitations of existing OBB detectors as well as to validate their performance after using the PIoU loss. In the future, we will extend PIoU to 3D rotated object detection. Our preliminary results show that PIoU can improve PointPillars~\cite{lang2019pointpillars} on KITTI val dataset~\cite{geiger2013vision} by 0.65, 0.64 and 2.0 AP for car, pedestrian and cyclist in moderate level, respectively.